\documentclass[11pt]{article}
\pdfoutput=1

\usepackage{graphicx} % more modern
\usepackage{subfigure} 

\usepackage{algorithm}
\usepackage{algorithmic}
\usepackage{amsmath,amsthm}
\usepackage{slashbox}
\usepackage{multirow}
\usepackage{amssymb}
\usepackage{hyperref}

\newcommand{\argmax}{\operatornamewithlimits{arg\,max}}
\begin{document}

%\twocolumn[
%\icmltitle{Greedy Multiple Instance Learning via Codebook Learning and Nearest Neighbor Voting}

% unless you've provided the [accepted] option to the icml2012
\title{{\bf Deep Learning with Nonparametric Clustering}}
\author{Gang Chen
%\thanks{SUNY at Buffalo}
}
\maketitle

\vskip 0.3in
%%%%%%%%% ABSTRACT

%%%%%%%%% ABSTRACT
\begin{abstract}
Clustering is an essential problem in machine learning and data mining. One vital factor that impacts clustering performance is how to learn or design the data representation (or features). Fortunately, recent advances in deep learning can learn unsupervised features effectively, and have yielded state of the art performance in many classification problems, such as character recognition, object recognition and document categorization. However, little attention has been paid to the potential of deep learning for unsupervised clustering problems. In this paper, we propose a deep belief network with nonparametric clustering. As an unsupervised method, our model first leverages the advantages of deep learning for feature representation and dimension reduction. Then, it performs nonparametric clustering under a maximum margin framework -- a discriminative clustering model and can be trained online efficiently in the code space. Lastly model parameters are refined in the deep belief network. Thus, this model can learn features for clustering and infer model complexity in an unified framework. The experimental results show the advantage of our approach over competitive baselines. 
%On the one hand, we exploit the unsupervised feature learning from deep learning approach; On the other hand, we leverage nonparametric clustering approach for clustering and model learning. 
\end{abstract}

% ------------------------------------------------------------------------------------------
\section{Introduction}
Clustering methods, such as k-means, Gaussian mixture model (GMM), spectral clustering and non-parametrical Bayesian methods, have been widely used in machine learning and data mining. Among various clustering methods, nonparametric Bayesian model is one of promising approaches for data clustering, because of its ability to infer the model complexity from the data automatically. To mine clusters or patterns from data, we can group them based on some notion of similarity. In general, calculating the clustering similarity is dependent on the features describing data. Thus, feature representation is vital for successful clustering. Just as common for other clustering methods, the presence of noisy and irrelevant features can degrade clustering performance, making feature representation an important factor in cluster analysis. Moreover, different features may be relevant or irrelevant in the high dimensional data, suggesting the need for feature learning. %In addition, the cost for computation in DPM is polynomially growing in dimensionality, making dimension reduction necessary in clustering analysis. 

Recent advances in deep learning \cite{hinton06a,Vincent10,Bengio12} have attracted great attention in dimension reduction \cite{Hinton06b,Weston08} and classification problems \cite{hinton06a,Larochelle12,Tang13}. The advantages of deep learning are that they give mappings which can capture meaningful structure information in the code space and introduce bias towards configurations of the parameter space that are helpful for unsupervised learning \cite{Erhan10}. More specifically, it learns the composition of multiple non-linear transformations (such as stacked restricted Boltzmann machines), with the purpose to yield more abstract and ultimately more useful representations \cite{Bengio12}. %In a sense, the success of deep learning lies on learned latent representations, which are helpful for supervised/unsupervised tasks \cite{Erhan10,Bengio12}. For example, the deep autoencoders \cite{Hinton06b} pre-trained with stacked restricted Boltzmann machines (RBMs), learn low-dimensional manifold by minimizing the reconstruction error to facilitate the classification and visualization of data. %And later, this unsupervised dimension reduction approach was developed into semi-supervised embedding \cite{Weston08} and supervised mapping \cite{Min10} scenarios. 
%Another popular deep structure model for feature learning is a deep belief network \cite{hinton06a}(or DBN), where the top layer is interpreted as an RBM and the lower layers as a directed sigmoid belief network. A further approach has been proposed to stack RBMs into a deep Boltzmann machine (DBM) \cite{Salakhutdinov09}, which is the fully generative model and can be trained by approximate maximum likelihood. The advantages of deep learning are that they give mappings which can capture meaningful structure information in the code space and introduce bias towards configurations of the parameter space that are helpful for unsupervised learning \cite{Erhan10}. More specifically, it learns the composition of multiple non-linear transformations (such as stacked restricted Boltzmann machines), with the purpose to yield more abstract and ultimately more useful representations \cite{Bengio12}. Moreover, the fine-tuning process can greatly improve performance after the greedy layer-wise unsupervised pre-training \cite{Salakhutdinov09}. %Moreover, a two-step process can greatly improve performance: unsupervised pre-training with greedy layer-wise learning, is then followed by a fine-tuning step in the algorithm \cite{Hinton06b,Salakhutdinov09,Tang13}. %
In addition, deep learning with gradient descent scales linearly in time and space with the number of train cases, which makes it possible to apply to large scale data sets \cite{Hinton06b}. 

Unfortunately, little work has been done to leverage the advantages of deep learning for unsupervised clustering problems. Moreover, unsupervised clustering also presents a challenge in the deep learning framework, compared to supervised methods in the final fine-tuning process. Another important research topic in clustering analysis is how to adapt model complexity for increasing volumes in the era of big data \cite{Rasmussen00,Blei05,Teh2010a}. However, most approaches are generative models and have restrictions on the prior base measures. %involves methods of adapting model complexity for increasing volumes in the era of big data \cite{Rasmussen00,Blei05,Teh2010a}. 
%It is a challenge to handle high dimension and large scale data sets in the era of big data. 

\begin{figure}
\centering
\includegraphics[trim = 60mm 36mm 100mm 32mm, clip, width=11.5cm]{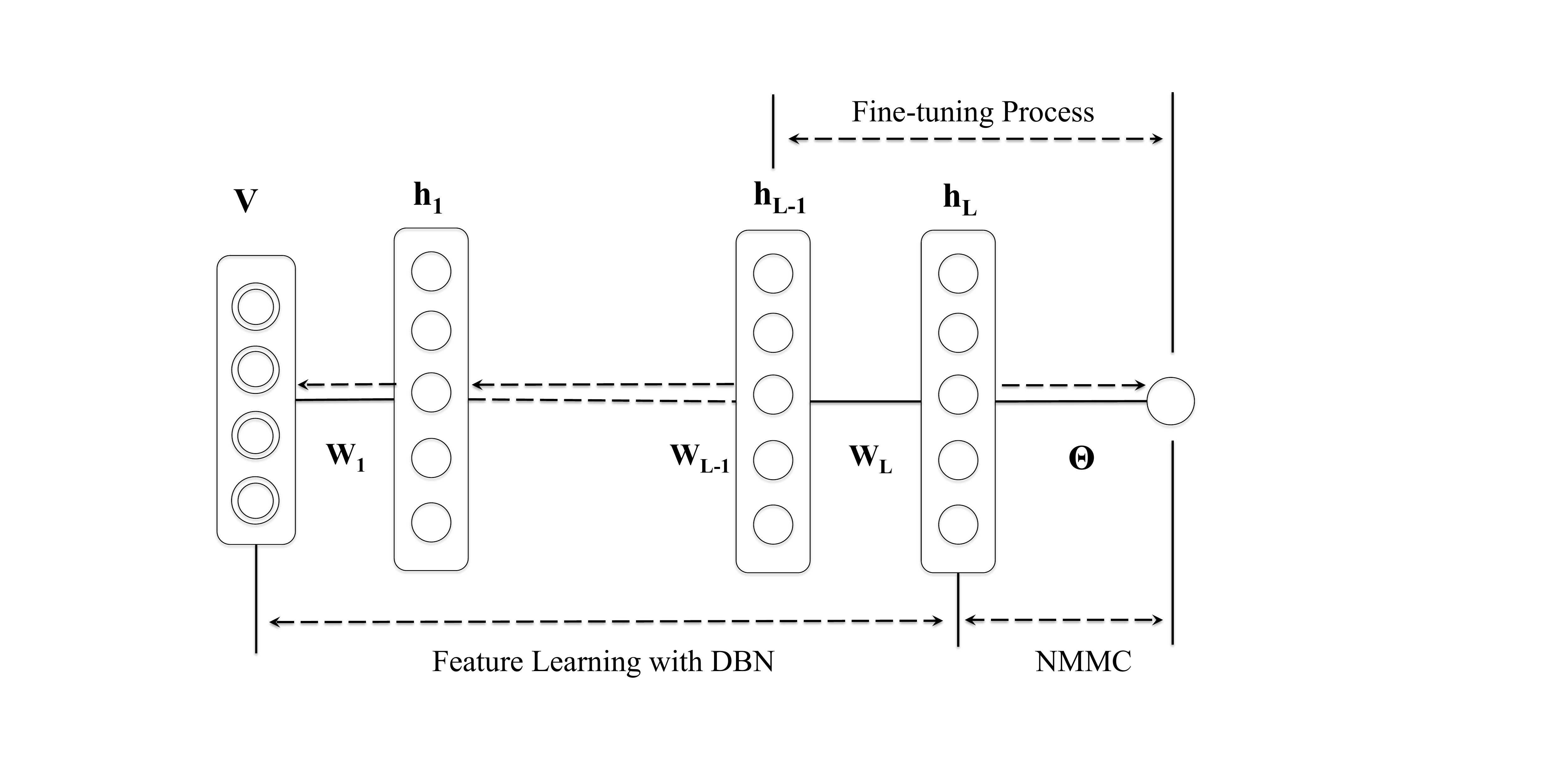}
\caption{In this DBN, $\textrm{L}$ indicates the total number of hidden layers, ${\bf W}_{i}$ is the weight between adjacent layers, for $i=\{1,..,\textrm{L}\}$ and $\mathbf{\Theta}$ is the weight for clustering learned with NMMC. This graph demonstrates 3 steps in our model: (1) Feature learning with deep belief network (DBN), with weights learned layer by layer as described above; (2) Perform clustering analysis with NMMC, which can assign a cluster label for each element in the data; (3) Update the model parameters with fine-tuning process (only for ${\bf W}_{\textrm{L}}$ and $\mathbf{\Theta}$).}
\label{fig:dbn}
\end{figure}

In this paper, we are interested in clustering problems and propose a deep belief network (DBN) with nonparametric clustering. This approach is an unsupervised clustering method, inspired by the advances in unsupervised feature learning with DBN, as well as nonparametric Bayesian models \cite{antoniak74,ferguson73,Blei05}. On the one hand, clustering performance depends heavily on data representation, which implies the need for feature learning in clustering. %which can be compensated with DBN for unsupervised feature learning. 
On the other hand, while the nonparametric Bayesian model can perform model selection and data clustering, it is intractable for non-conjugate prior; furthermore, it may not perform well on high-dimensional data, especially in terms of space and time complexity. Thus, we propose the deep learning with nonparametric maximum margin model for clustering analysis. Essentially, we first pre-train DBN for feature learning and dimension reduction. Then, we will learn the clustering weights discriminatively with nonparametric maximum margin clustering (NMMC), which can be updated online efficiently. Finally, we fine-tune the model parameters in the deep belief network. Refer to Fig. (\ref{fig:dbn}) for visual understanding to our model. Hence, our framework can handle high-dimensional input features with nonlinear mapping, and cluster large scale data sets with model selection using the online nonparametric clustering method. 

Our contributions can be mainly summarized as: (1) leveraging unsupervised feature learning with DBN for clustering analysis; (2) a discriminative approach for nonparametric clustering under maximum margin framework. %We will release the reproducible code after acceptance. 
The experimental results show advantages of our model over competitive baselines. %We will release the reproducible code after acceptance. 
%with the help from autoencoder embedding which preserve meaningful structure information. , our method perform clustering in the nonlinear embedding space, 

%Our method is motived by the following two reasons: (1) Autoencoder pre-trained with restricted Boltzmann machine (RBM) can capture meaningful structure from high dimensional input features, which is useful for visualization and classification problems; (2) Nonparametric methods can automatically infer model size or complexity from the data. 

%Our method with 3 steps:
%(1) nonlinear mapping with pre-trained restricted Boltzmann machine;
%(2) nonparametric maximum margin clustering in the embedding space;
%(3) fine-tune the model given the clustering labels.
% ---------------------------------------------------------------------------------------
\section{Related work}
Clustering has been an interesting research topic for decades, including a wide range of techniques, such as generative/discriminative and parametric/nonparametric approaches. As an discriminative method, maximum margin clustering (MMC) treats the label of each instance as a latent variable and uses SVM for clustering with large margins. However, they \cite{Ben-Hur01,Xu05} either cannot learn parameters online efficiently or need to define the number of clusters like other clustering approaches, such as k-means, Gaussian mixture model (GMM) and spectral clustering. 
%Recently developments in nonparametric models ignite the hope for unsupervised clustering and model selection. The related work to our nonparametric maximum margin clustering (NMCC) is Dirichlet process mixture (DPM) \cite{antoniak74,ferguson73,Blei05}. However, DPM is a generative model, while our method is a discriminative approach for clustering. 
%However, majority nonparametric methods are generative models, which is hard to unify them with RBM in the fine-tuning step. Recent, Tang et. al. proposed a supervised classification method using deep learning with SVM \cite{Tang13}, which can fine-tune the lower layers by backpropagating the gradients from top layer linear SVM. 
%Instead, we advocate discriminative clustering in the embedding space, and use maximum margin clustering for label inference. Given the number of clusters and the labeled instances, we finally fine-tune our model with backpropagation algorithm. 
Considering the weakness of parametric models mentioned above, many nonparametric methods \cite{Blei05,Kurihara07,Hannah11,Knowles12} have been proposed to handle the model complexity problems. One of the widely used nonparametric models for clustering is Dirichlet process mixture (DPM) \cite{antoniak74,ferguson73}. DPM can learn the number of mixture components without specified in advance, which can grow as new data come in. However, the behavior of the model is sensitive to the choice of prior base measure $G_0$. In addition, DPM of Gaussians need to calculate mean and covariance for each component, and update covariance with Cholesky decomposition, which may lead to high space and time complexity in high-dimensional data. %Furthermore, as a generative model, it is intractable for non-conjugate prior. %Thus, we consider a discriminative approach for nonparametric clustering (or NMMC). In order to handle high-dimensional features, we use deep learning for feature learning and NMCC for clustering.
Unsupervised feature learning with deep structures was first proposed in \cite{Hinton06b} for dimension reduction. Later, this unsupervised approach was developed into semi-supervised embedding \cite{Weston08} and supervised mapping \cite{Min10} scenarios. Many other supervised approaches also exploit deep learning for feature extraction and then learn a discriminative classifier with objectives, e.g., square loss \cite{Hinton06b}, logistic regression \cite{Larochelle12} or support vector machine (SVM) \cite{Krizhevsky12,Tang13} for classification in the code space. The success behind deep learning is that it can learn useful information for data visualization and classification \cite{Erhan10,Bengio12}. Thus, it is desirable to leverage deep learning for clustering analysis, because the performance for clustering depends heavily on data representation. Unfortunately, little attention has been paid to leveraging deep learning for unsupervised clustering problems. 
%The approaches in \cite{hinton06a,Larochelle12} is close to our method. The basic ideas are to exploit RBM or multilayer RBMs for supervised classification problems.  
% Another hot topic is to exploit RBM for classification problems. 

A recent interesting approach is the implicit mixture of RBMs \cite{Nair08}. Instead of modeling each component with Gaussian distribution, it models each component with RBM. It is formulated as a third-order Boltzmann machine with cluster label as the hidden variable for each instance. However, it also requires the number of clusters specified as input. 
 %Recently, Tang proposed a deep learning approach with SVM \cite{Tang13}. This method is supervised approach with label given for each training data, and also it requires the number of classes as input. 

In this paper, we are interested in deep learning for unsupervised clustering problems. In our framework, we take advantage of deep learning for representation learning, which is helpful for clustering analysis. Moreover, we take an discriminative approach, namely nonparametric maximum margin clustering to infer model complexity online, without the prior measure assumption as DPM. %we think the number of clusters and labels for all instances are latent variables. On the one hand, for unsupervised clustering, we need to solve the model selection problem. On the other hand, we should fine-tune the RBM parameters after we learned the number of clusters. Thus, it is a challenge, compared to supervised classification problems with label information. 

\section{Deep learning with nonparametric maximum margin clustering}
In this section, we will first review RBM and DBN for feature learning. Then, we will introduce nonparametric maximum margin clustering (NMMC) method given the feature learned from DBN. Finally, we will fine-tune our model given the clustering labels for the data. 
\subsection{Feature learning with deep belief network}
Assume that we have a training set $\mathcal{D} = \{{\bf v}_i\}_{i=1}^N$, where ${\bf v}_{i} \in \mathbb{R}^d$. An RBM with $n$ hidden units is a parametric model of the joint distribution between a layer of hidden variables ${\bf h} = (h_1,...,h_n)$ and the observations ${\bf v} = (v_1, ..., v_d)$. The RBM joint likelihood takes the form:
\begin{equation}\label{eq:eq1}
p({\bf v},{\bf h}) \propto e^{-E({\bf v},{\bf h})}
\end{equation}
where the energy function is 
\begin{equation}\label{eq:eq2}
E({\bf v},{\bf h}) = -{\bf h}^T{\bf W}_1{\bf v} - {\bf b}^T {\bf v} - {\bf c}^T{\bf h} 
\end{equation}
% with parameters $\Theta_{1} = \{{\bf W}^1, {\bf b}, {\bf c}\}$. 
And we can compute the following conditional likelihood:
\begin{subequations}\label{eq:eq3}
\begin{align}
        & p({\bf v} | {\bf h}) = \prod_{i} p(v_{i} | {\bf h})\\
        & p(v_i=1 | {\bf h}) = \mathrm{logistic}(b_{i} + \sum_{j} W_1(i,j) h_j)\\
        & p(h_i =1 | {\bf v}) = \mathrm{logistic}(c_{i} + \sum_{j} W_1(j, i)v_j)
\end{align}
\end{subequations}
where $\mathrm{logistic}(x)  = 1/(1+e^{-x})$. To learn RBM parameters, we need to optimize the negative log likelihood $-\mathrm{log}p({\bf v})$ on training data $\mathcal{D}$, the parameters updating can be calculated with a efficient stochastic descent method, namely contrastive divergence (CD) \cite{hinton06a}.% to the negative log likelihood.
%$\{{\bf W}_{i}\}_{i=1}^L$

A Deep Belief Network (DBN) is composed of stacked RBMs \cite{Hinton06b} learned layer by layer greedily, where the top layer is an RBM and the lower layers can be interpreted as a directed sigmoid belief network \cite{Bengio12}, shown in Fig. (\ref{fig:dbn}). Suppose the DBN used here has $\textrm{L}$ layers, and the weight for each layer is indicated as ${\bf W}_{i}$ for $i=\{1,..,\textrm{L}\}$. Specifically, we think RBM is a 1-layer DBN, with weight ${\bf W}_1$. Thus, DBN can learn parametric nonlinear mapping from input ${\bf v}$ to output ${\bf x}$, $f: {\bf v} \rightarrow {\bf x}$. For example, for 1-layer DBN, we have ${\bf x} = \textrm{logistic}({\bf W_{1}}^T {\bf v} +{\bf c})$. After we learn the representation for the data, we use NMCC for clustering analysis to model the data distribution. 
%$x = f(v; \{{\bf W}_{i}\}_{i=1}^L)$
%However, there is some empirical evidence that the second layer of the DBN tends to display more invariance than the first layer \cite{Erhan10}.

\subsection{Nonparametric maximum margin clustering}
Nonparametric maximum margin clustering (NMMC) is a discriminative clustering model for clustering analysis. Given the nonlinear mapping with DBN, we can first map the original training data $\mathcal{D} = \{{\bf v}_i\}_{i=1}^N$ into codes $\mathcal{X} = \{{\bf x}_i\}_{i=1}^N$ in the embedding space. Then, with ${\bf \mathcal{X}} = \{{\bf x}_i\}_{i=1}^N$ and its the cluster indicators ${\bf z} = \{z_i\}_{i=1}^N$, we propose the following conditional probability for nonparametric clustering:
\begin{flalign}
P({\bf z}, \{\boldsymbol{\theta}_{k}\}_{k=1}^K | \mathcal{X})   \propto  p({\bf z}) \bigg[ \prod_{i=1}^N p({\bf x}_i | \boldsymbol{\theta}_{z_{i}})\bigg]  \prod_{k=1}^K p(\boldsymbol{\theta}_{k})
 \label{eq:dpmdis} 
%= & \quad \pi_{k} p(x_{i} | \theta_{k}) \label{eq:gen4}
\end{flalign}
where $K$ is the number of clusters, $p({\bf x}_i | \boldsymbol{\theta}_{z_{i}})$ is the likelihood term defined in Sec. \ref{sec:gibbs} and $p(\boldsymbol{\theta}_{k})$ can be thought as the Gaussian prior for $k=[1,...,K]$. Note that the prior $p(\boldsymbol{\theta}_{k})$ will be used in the maximum margin learning in Eq. (\ref{eq:onlinesvm}). 
$p({\bf z}) = \frac{\Gamma (\alpha) \prod_{k=1}^K \Gamma(n_k + \alpha/K)}{\Gamma(n+\alpha) \Gamma(\alpha/K)^K}$ is the symmetric Dirichlet prior, where $n_k$ is the number of element in the cluster $k$, and $\alpha$ is the concentration parameter.

Recall that Dirichlet process mixture (DPM) \cite{antoniak74,ferguson73} is the widely used nonparametric Bayesian approach for clustering analysis and model learning, specified with DP prior measure $G_{0}$ and $\alpha$. %A DPM model can be constructed as a limit of a parametric mixture model \cite{Neal00,Rasmussen00,Blei05} with the DP prior $G_{0}$ and $\alpha$. 
As a joint likelihood model, it has to model $p(\mathcal{X})$, which is intractable for non-conjugate prior. The essential difference between our model and DPM is that we maximize a conditional probability, instead of joint probability as in DPM \cite{Kurihara07}. Moreover, our approach is a discriminative clustering model with component parameters learned under maximum margin framework. %, which has no assumption on prior measure for $\{\boldsymbol{\theta}_{k}\}_{k=1}^K$. In other words, we maximize a conditional probability, instead of joint probability as in DPM \cite{Kurihara07}. 

To maximizing the objective function in Eq. (\ref{eq:dpmdis}), we hope the higher within-cluster correlation and lower correlation between different clusters. Given ${\bf z}$, we will need to learn $\{\boldsymbol{\theta}_{k}\}_{k=1}^K$ to keep each cluster as compact as possible, which in turn will help infer better $K$. In other words, to keep the objective climbing, we need higher likelihood $p({\bf x}_i | \boldsymbol{\theta}_{z_{i}})$ with higher correlation within-cluster, which can be addressed with discriminative clustering. Given the component parameters, $\{\boldsymbol{\theta}_{k}\}_{k=1}^K$, we need to decide the label for each element for better $K$. For each round (on the instance level), we use Gibbs sampling to infer $z_{i}$ for each instance ${\bf x}_i$, which in turn can be used to estimate $\{\boldsymbol{\theta}_{k}\}_{k=1}^K$ with online maximum margin learning. For each iteration (on the whole dataset), we also update $\alpha$ with adaptive rejection sampling \cite{Neal00}.

\subsubsection{Gibbs sampling}\label{sec:gibbs}
%Before discussing Gibbs sampling for DPM, we first see the case of finite parametric mixture model (FMM). In a FMM, 
Given the data points ${\bf \mathcal{X}} = \{{\bf x}_i\}_{i=1}^N$ and its the cluster indicators ${\bf z} = \{z_i\}_{i=1}^N$, the Gibbs sampling involves iterations that alternately draw samples from conditional probability while keeping other variables fixed. %one of the following variables while keeping others fixed: the cluster indicators ${\bf z}$, the cluster parameters $\{\theta_k\}_{k=1}^K$ and the mixture weights ${\boldsymbol \pi}$. %The first step towards Gibbs sampling is to derive the conditional posterior distributions for these variables. The hyperparameter $\alpha$ and $\beta$ are assumed known in this process. By exploiting the Markov properties of the FMM and employing the Bayes rule, these distributions will be simplified to great extents. 
For each indicator variable $z_i$, we can derive its conditional posterior as follows:
\begin{flalign}
&\quad p(z_{i} = k | {\bf z_{-i}}, {\bf x}_i, \{\boldsymbol{\theta}_{k}\}_{k=1}^K, \alpha, \lambda) \\
=&\quad p(z_{i} = k | {\bf x}_i, {\bf z_{-i}}, \{\boldsymbol{\theta}_{k}\}_{k=1}^K) \label{eq:gen1}  \\
\propto & \quad p(z_{i} = k | {\bf z_{-i}}, \{\boldsymbol{\theta}_{k}\}_{k=1}^K)p({\bf x}_i| z_{i} = k, \{\boldsymbol{\theta}_{k}\}_{k=1}^K) \label{eq:gen2}\\
= & \quad p(z_{i} = k | {\bf z_{-i}}, \alpha)p({\bf x}_i |\boldsymbol{\theta}_{k}) % \label{eq:gen3} 
%= & \quad \pi_{k} p(x_{i} | \theta_{k}) \label{eq:gen4}
\label{eq:gibbs}
\end{flalign}
where the subscript $-i$ indicates all indices except for $i$, $p(z_{i} = k | {\bf z_{-i}}, \alpha)$ is determined by Chinese restaurant process, and $p({\bf x}_i |\boldsymbol{\theta}_{k}) \label{eq:gen3}$ is the likelihood for the current observation ${\bf x}_i$. For DPM, we need to maximize the conditional posterior to compute $\boldsymbol{\theta}_k$, which depends on observations belonging to this cluster and prior $G_{0}$.
In our conditional likelihood model, we define the following likelihood for instance ${\bf x}_i$%we replace the generative model in DPM with our discriminative SVM classifier. We define the following likelihood for instance ${\bf x}_i$:
\begin{equation}
%p(z_{i} = k | {\bf x}_i, \{\theta_{k}\}_{k=1}^K) \propto exp({\bf x}_{i}^T{\bf \theta}_k - \lambda || {\bf \theta}_k||^2)
p({\bf x}_i| \boldsymbol{\theta}_{k}) \propto exp({\bf x}_{i}^T \boldsymbol{\theta}_k - \lambda || \boldsymbol{\theta}_k||^2)
\label{eq:condprob}
\end{equation}
where $\lambda$ is a regularization constant to control weights between the two terms above. %SVM classifiers never really output an actual probability. The output of an SVM classifier is the distance of the test instance to the separating hyperplane in feature space (this is called the decision value). 
By default, the prediction function should be proportional to $\argmax_{k}({\bf x}_i^T\boldsymbol{\theta}_k)$, for $k \in [1,K]$. In other words, higher correlation between ${\bf x}_i$ and $\boldsymbol{\theta}_k$ indicates higher probability that ${\bf x}_i$ belongs to cluster $k$, which further leads to higher objective in Eq. (\ref{eq:dpmdis}).
In our likelihood definition, we also subtract $\lambda || \boldsymbol{\theta}_k||^2$ in Eq. (\ref{eq:condprob}), which can keep the maximum margin beneficial properties in the model to separate clusters as far away as possible. %Moreover, it can get rid of trivial clustering results \cite{Hoai13}. Note that the seminal work \cite{Platt99} basically fits a sigmoid function over SVM decision values to scale it to the range of $[0, 1]$, which can then be interpreted as a kind of probability. %Similar techniques can be applied on any type of classifier that produces a real-valued output. 
%Compared to \cite{Platt99}, we maximize $({\bf x}_{i}^T{\bf \theta}_k)$ and minimize $\lambda|| {\bf \theta}_k||^2$ simultaneously in Eq. (\ref{eq:condprob}), so our method can keep larger margins between clusters. 
Another understanding for the above likelihood is that Eq. (\ref{eq:condprob}) satisfies the general form of exponential families, which are functions solely of the chosen sufficient statistics \cite{Sudderth06}. Thus, such probability assumption in Eq. (\ref{eq:condprob}) make it general to real applications. 

Plug Eq. (\ref{eq:condprob}) into Eq. (\ref{eq:gibbs}), we get the final Gibbs sampling strategy for our model%maximum margin clustering model:
%\begin{flalign}
%& \quad p(z_{i} = k | {\bf z_{-i}}, {\bf x}_i, \{\theta_{k}\}_{k=1}^K, \alpha, \beta) \nonumber \\
%& \propto p(z_{i} = k | {\bf z_{-i}}, \alpha) exp({\bf x}_{i}^T{\bf \theta}_k - \lambda || {\bf \theta}_k||^2)
%\label{eq:finalgibbs}
%\end{flalign}
\begin{flalign}
& \quad p(z_{i} = k | {\bf z_{-i}}, {\bf x}_i, \{\boldsymbol{\theta}_{k}\}_{k=1}^K, \alpha, \lambda)  \nonumber \\
& \propto  p(z_{i} = k | {\bf z_{-i}}, \alpha) exp({\bf x}_{i}^T\boldsymbol{\theta}_k - \lambda || \boldsymbol{\theta}_k||^2)
\label{eq:finalgibbs}
\end{flalign}
We will introduce online maximum margin learning for component parameters $\{\boldsymbol{\theta}_{k}\}_{k=1}^K$ in Sec \ref{sec:mml}.
For the newly created cluster, we assume $\boldsymbol{\theta}_{K+1}$ is sampled from multivariate t-distribution.
%one approach \cite{Hoai13} generates $\theta_{K+1}$ that is perpendicular to all the previous $\theta_{k}$, for $k \in[1, K]$. In our paper, we assume $\theta_{K+1}$ is sampled from multivariate t-distribution.

\subsubsection{Online maximum margin learning}\label{sec:mml}
\label{mmc}
%In this part, we introduce to learn components parameters $\{\theta_{k}\}_{k=1}^K$ with discriminative model.
We follow the passive aggressive algorithm (PA) \cite{Crammer2006} below in order to learn component parameters in our discriminative model with maximum margins \cite{Vapnik1995}. %Basically, the online algorithm is a kind of perception algorithm. It observes instances in a sequential manner, and then infers an outcome with the current model. If the prediction mismatches its feedback, then the online algorithm update its model, presumably improving the chances of making an accurate prediction on subsequent rounds. 

We denote the instance presented to the algorithm on round $t$ by ${\bf x}_{t} \in \mathbb{R}^n$, which is associated with a unique label $z_{t} \in [1, K]$. Note that the label $z_{t}$ is determined by the above Gibbs sampling algorithm in Eq. (\ref{eq:finalgibbs}). We shall define $\mathbf{\Theta} = [\boldsymbol{\theta}_1, ..., \boldsymbol{\theta}_K]$ a parameter vector by concatenating all the parameters $\{\boldsymbol{\theta}_{k}\}_{k=1}^K$ (that means $\mathbf{\Theta}^{z_t}$ is $z_t$-th block in $\mathbf{\Theta}$, or says $\mathbf{\Theta}^{z_t}$ = $\boldsymbol{\theta}_{z_t}$), and ${\bf \Phi}({\bf x}_t, z_t)$ is a feature vector relating input ${\bf x}_t$ and output $z_t$, which is composed of $K$ blocks, and all blocks but the $z_t$-th are set to be the zero vector while the $z_t$-th block is set to be ${\bf x}_t$.
We denote by $\mathbf{\Theta}_{t}$ the weight vector used by the algorithm on round $t$,
and refer to the term  $\gamma({\mathbf{\Theta}_{t}}; ({\bf x}_t, z_t)) =  {\mathbf{\Theta}_{t}}\cdot {\bf \Phi}(\textbf{x}_{t}, z_{t})- {\mathbf{\Theta}_{t}}\cdot {\bf \Phi}({\bf x}_{t}, \hat{z_{t}})$ %${\bf w_{t}}\cdot {\bf \Phi}({\bf x}_{t}, z_{t})- {\bf w_{t}}\cdot {\bf \Phi}({\bf x}_{t}, \hat{z_{t}})$ 
 as the (signed) margin attained on round $t$. 
%Note that the principle of maximum margin svm is to achieve a margin of at least 1 as often as possible, otherwise it suffers an instantaneous loss if it is a wrong prediction. 
In this paper, we use the hinge-loss function, which is defined by the following,
\begin{equation}\label{eq:mmc}
\begin{split}
 \ell(\mathbf{\Theta}; ({\bf x}_{t},z_{t})) = \left\{ \begin{array}{ll}
 0 &\mbox{ if $\gamma({\mathbf{\Theta}_{t}}; ({\bf x}_t, z_t))  \ge 1$} \\
  1-\gamma({\mathbf{\Theta}_{t}}; ({\bf x}_t, z_t))  &\mbox{ otherwise}
       \end{array} \right.
\end{split}       
\end{equation}
% where $\gamma({\bf w_{t}}; ({\bf x}_t, z_t)) =  {\bf w_{t}}\cdot {\bf \Phi}(\textbf{x}_{t}, z_{t})- {\bf w_{t}}\cdot {\bf \Phi}({\bf x}_{t}, \hat{z_{t}})$ is the margin attained by the algorithm on round $t$.
 Following the passive aggressive (PA) algorithm \cite{Crammer2006}, we optimize the objective function:
 \begin{equation}\label{eq:onlinesvm}
 \begin{split}
 & \mathbf{\Theta}_{t+1} = \underset{\mathbf{\Theta}}{\arg\min} \frac{1}{2} ||\mathbf{\Theta} - \mathbf{\Theta}_{t}||^2+ C\xi \\
 & \qquad s.t.  \ \ell(\mathbf{\Theta}; (\mathbf{x}_{t},z_{t})) \le \xi
 \end{split}
 \end{equation}
 where the $l_2$ norm of $\mathbf{\Theta}$ on the right hand size can be thought as Gaussian prior in Eq. (\ref{eq:dpmdis}). If there's loss, then the updates of PA-1 has the following closed form
 \begin{equation}
\begin{array}{l}
\displaystyle   \mathbf{\Theta}_{t+1}^{z_t} =  \mathbf{\Theta}_{t}^{z_t}  + \tau_{t} {\bf x}_{t},  \\
\displaystyle   \mathbf{\Theta}_{t+1}^{\hat{z_t}} = \mathbf{\Theta}_{t}^{\hat{z_t}}  - \tau_{t} {\bf x}_{t}, 
\end{array} 
\label{eq:update}
\end{equation}
where $\hat{z_t}$ is the label prediction for ${\bf x}_{t}$, and $\tau_{t} = \min\{C, \frac{\ell(\mathbf{\Theta}_t; (\mathbf{x}_t, z_t))}{||\mathbf{x}_{t}||^2}\}$.
Note that the Gibbs sampling step can decide the indicator variable $z_{t}$ for ${\bf x}_t$. Given the cluster label (the ground truth assignment) for ${\bf x}_t$, we update our parameter $\mathbf{\Theta}$ using the above Eq. (\ref{eq:update}). For convergence analysis and time complexity, refer to \cite{Crammer2006}.
 
%{\bf parameter space analysis:} If the data dimension is $d$, and the current cluster number is $K$, then $\mathbf{\Theta}$ need $d\times K$ in our model. Yet for the DPM model, if we assume a Gaussian distribution, we need to update and store $d^2\times K$ for the covariance matrix, and that is computationally expensive for high dimensional data. Even for the diagonal covariance matrix, it still requires $2d\times K$ to store both mean and covariance.
%{\bf classification}:
% -----------------------------    ---------------------------------
\subsection{Fine-tuning the model}
Having determined the number of clusters and labels for all training data, we can take the fine-tuning process to refine the DBN parameters. Note that the objective function in Eq. (\ref{eq:onlinesvm}) takes the $l_1$ hinge loss as in \cite{Tang13}. Thus, one possible way is that we can take the sub-gradient and backpropagate the error to update DBN parameters. %In our fine-tuning process, we do just update the penultimate layer weights in the stacked deep neural network. 
In our approach, we employ another method and only update the top layer weights ${\bf W}_\textrm{L}$ and $\mathbf{\Theta}$ in the deep structures. %We settle for this strategy because of the exponential number of possible configurations of hyperparameters to be updated with gradient descent.%We do not backpropagate the error back in order to avoid overfitting. 
This fine-tuning process is inspired by the classification RBM \cite{Larochelle12} for model refining. Basically, we assume the top DBN layer weight ${\bf W}_\textrm{L}$ and SVM weight $\mathbf{\Theta}$ can be combined into a classification RBM as in \cite{Larochelle12}  by maximizing the joint likelihood $p({\bf x}, z)$ after we infer the cluster labels for all instances with NMMC. Note that there is mapping from SVM's scores to probabilistic outputs with logistic function \cite{Platt99}, which can maintain label consistency between the SVM classifier and the softmax function. Thus, the SVM weight $\mathbf{\Theta}$ can be used to initialize the weight of the softmax function in the classification RBM. After the fine-tuning process, we can $\max_{z}p(z| {\bf v})$ for $z \in [1, K]$ to label the unknown data ${\bf v}$. For 1-layer DBN, we can get the following classification probability:
%{\bf classification}
%where the energy function is 
%\begin{equation}\label{eq:eq2}
%E(y,{\bf x},{\bf h}) = -{\bf h}^T{\bf W}{\bf x} - {\bf b}^T {\bf x} - {\bf c}^T{\bf h} - {\bf d}^T {\bf y} - {\bf h}^T{\bf U} {\bf y}
%\end{equation}
%with parameters $\Theta = \{{\bf W}, {\bf b}, {\bf c}, {\bf d}, {\bf U}\}$ and ${\bf y} = (1_{y=i})_{i=1}^C$ for $C$ classes.
%And we can compute the following conditional likelihood:
%\begin{subequations}\label{eq:eq3}
%\begin{align}
%        & p({\bf x} | {\bf h}) = \prod_{i} p(x_{i} | {\bf h})\\
%        & p(x_i=1 | {\bf h}) = \mathrm{logistic}(b_{i} + \sum_{j} W_{ji}h_j)\\
%        & p(y | {\bf h}) = \frac{e^{d_{y}+\sum_{j} U_{jy}h_{j}} } {\sum_{y^*} e^{d_{y*}+\sum_{j} U_{jy*}h_{j}} }
%\end{align}
%\end{subequations}
%where $\mathrm{logistic}(x)  = 1/(1+e^{-x})$. To learn RBM parameters, we need to optimize the joint likelihood $p(y, {\bf x})$ on training data $\mathcal{D}$. Note that it is intractable to compute $p(y, {\bf x})$, because it needs to model $p(x)$. Fortunately,  Hinton proposed a efficient stochastic descent method, namely contrastive divergence (CD) \cite{Hinton06a} to maximize the joint likelihood.
%
%For classification problem, we need to compute the conditional probability for $p(y| {\bf x})$. As shown in \cite{Salakhutdinov07}, this conditional distribution has explicit formula and can be calculated exactly, by writing it as follows:
\begin{equation}\label{eq:eq4}
p(z | {\bf v})  = \frac{e^{d_{z}} \prod_{j=1}^n \big(1+e^{c_{j}+\mathbf{\Theta}_{jz} + \sum_{i} W_1(i, j)v_i} \big)}{\sum_{z^*} e^{d_{z^*}} \prod_{j=1}^n \big(1+e^{c_{j}+\mathbf{\Theta}_{jz^*} + \sum_{i} W_1(i, j)v_i} \big)}
\end{equation}
where $d_z$ for $z \in [1, K]$ is the bias of clustering labels, and $c_j$ for $j\in [1, n]$ are biases of the hidden units. Note that $\mathbf{\Theta}$ has been reshaped into $n \times K$ matrix before updating in the fine-tuning process. For the deep neural network with more than one layer, we first project ${\bf v}$ into the coding space ${\bf x}$, then use the above equation for classification.

In our algorithm, we only fine-tune in the top layer because of the following reasons: (1) the objective function in Eq. (\ref{eq:dpmdis}) with deep feature learning is non-convex, which can be easily trapped into local minimum with L-BFGS \cite{Hinton06b}; (2) if there was clustering error in the top layer, it could be easily propagated in the backpropagation stage; (3) To only update the top layer can effectively handle the overfitting problem. 
% ------------------------------------------------------------------------
\section{Experimental Results}
In order to analyze our model, we performed clustering analysis on two types of data: images and documents, and compared our results to competitive baselines. For all experiments, including pre-training and fine-tuning, we set the learning rate as 0.1, the maximum epoch to be 100, and used CD-1 to learn the weights and biases in the deep belief network. %layer by layer to pre-train the $\textrm{L}$ layers DBN and fine-tune the top layer model. In the fine-tuning process, only the parameters of the top layer were optimized.  
We used the adjusted Rand Index \cite{Hubert85,Rand71} to evaluate all the clustering results.
\begin{figure*}[t!]
  \centering
     \begin{tabular}{cc}
     \includegraphics[trim = 65mm 100mm 65mm 98mm, clip, width=6.0cm]{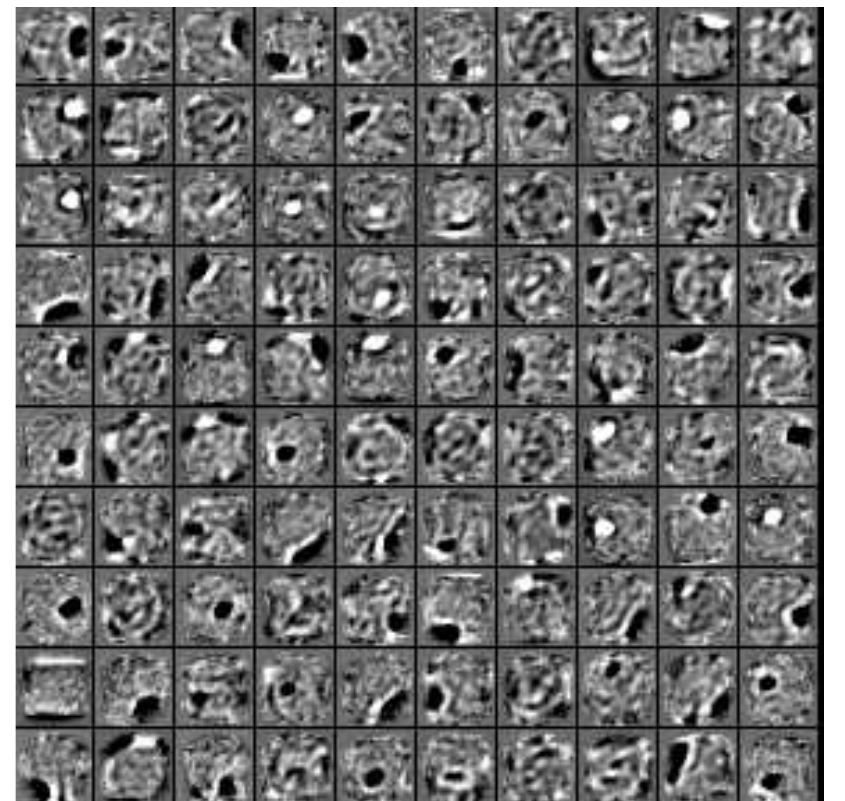} &
     \includegraphics[trim = 65mm 100mm 65mm 98mm, clip, width=6.0cm]{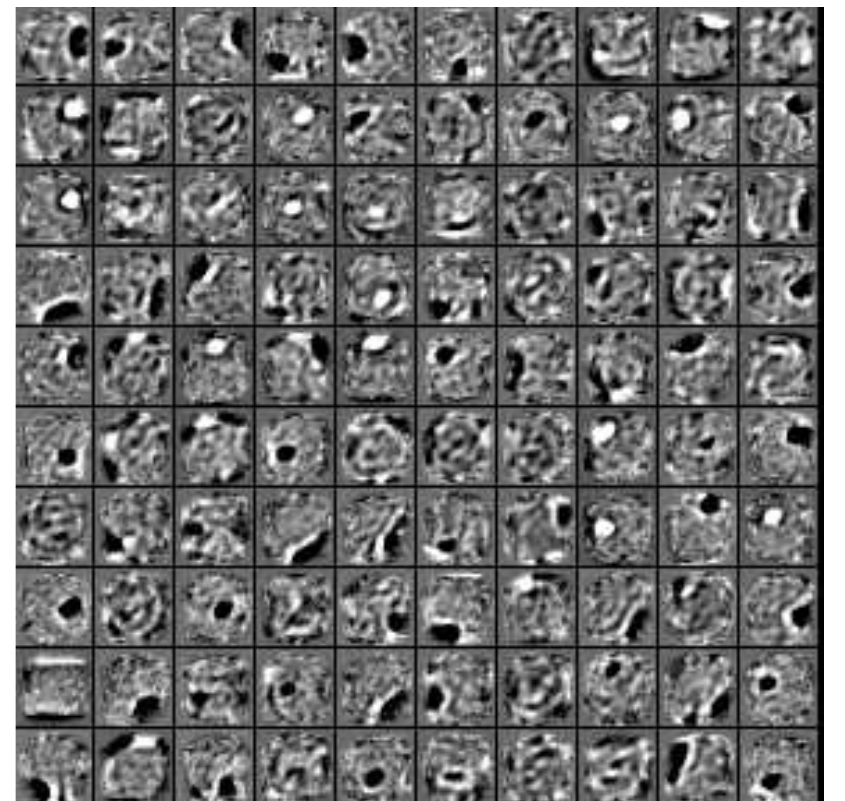}  \\ 
     (a) & (b)
     \end{tabular}
  \caption{The visualization of learned weights in the pre-training and fine-tuning stages respectively with 1-layer DBN for $n=100$ on the MNIST dataset.}
  \label{fig:weights}
\end{figure*}

\begin{figure*}[t!]
  \begin{center}
     \begin{tabular}{cc}
     \includegraphics[trim = 37mm 83mm 41mm 83mm, clip, width=6.7cm]{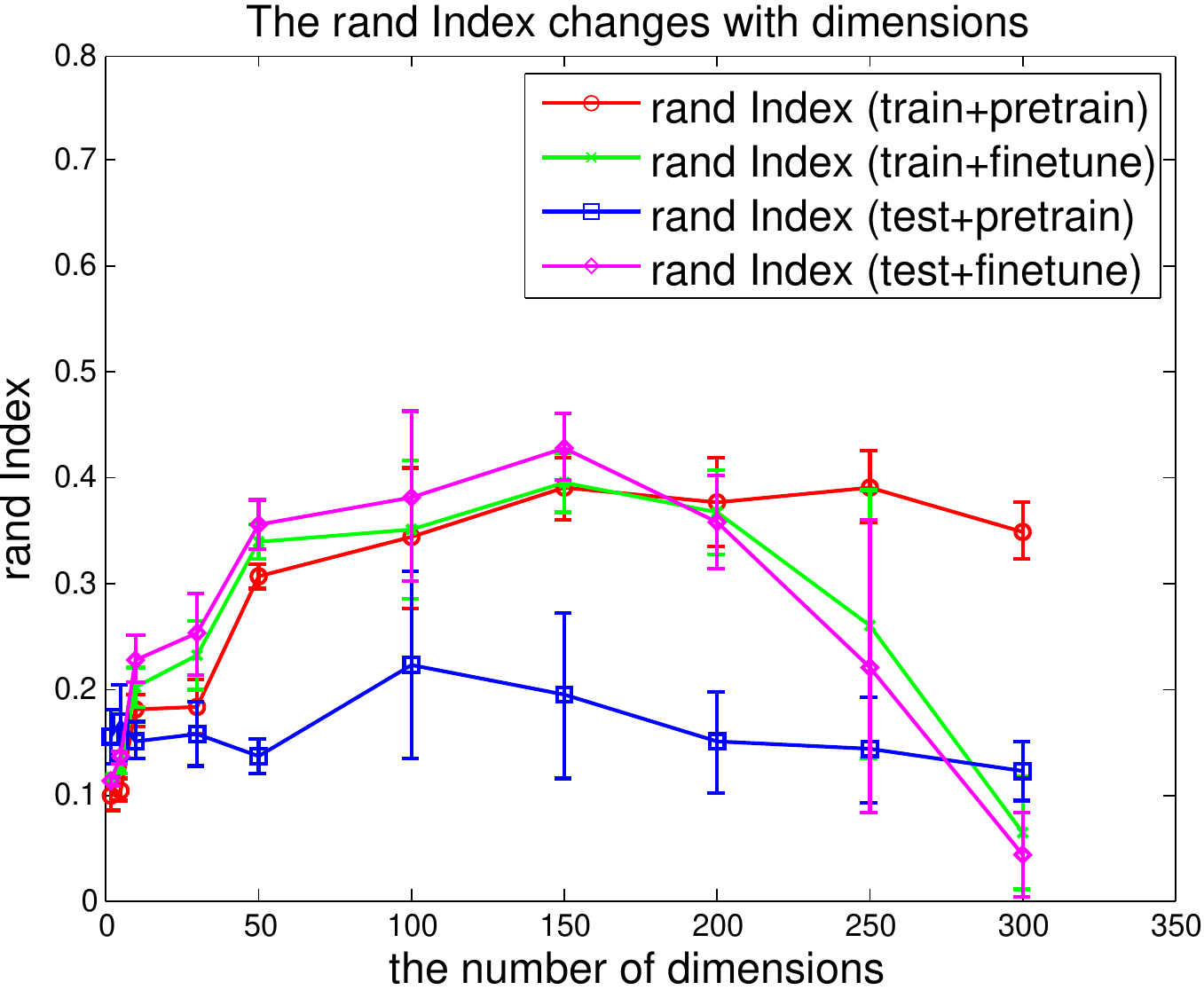} &
     \includegraphics[trim = 37mm 83mm 43mm 83mm, clip, width=6.6cm]{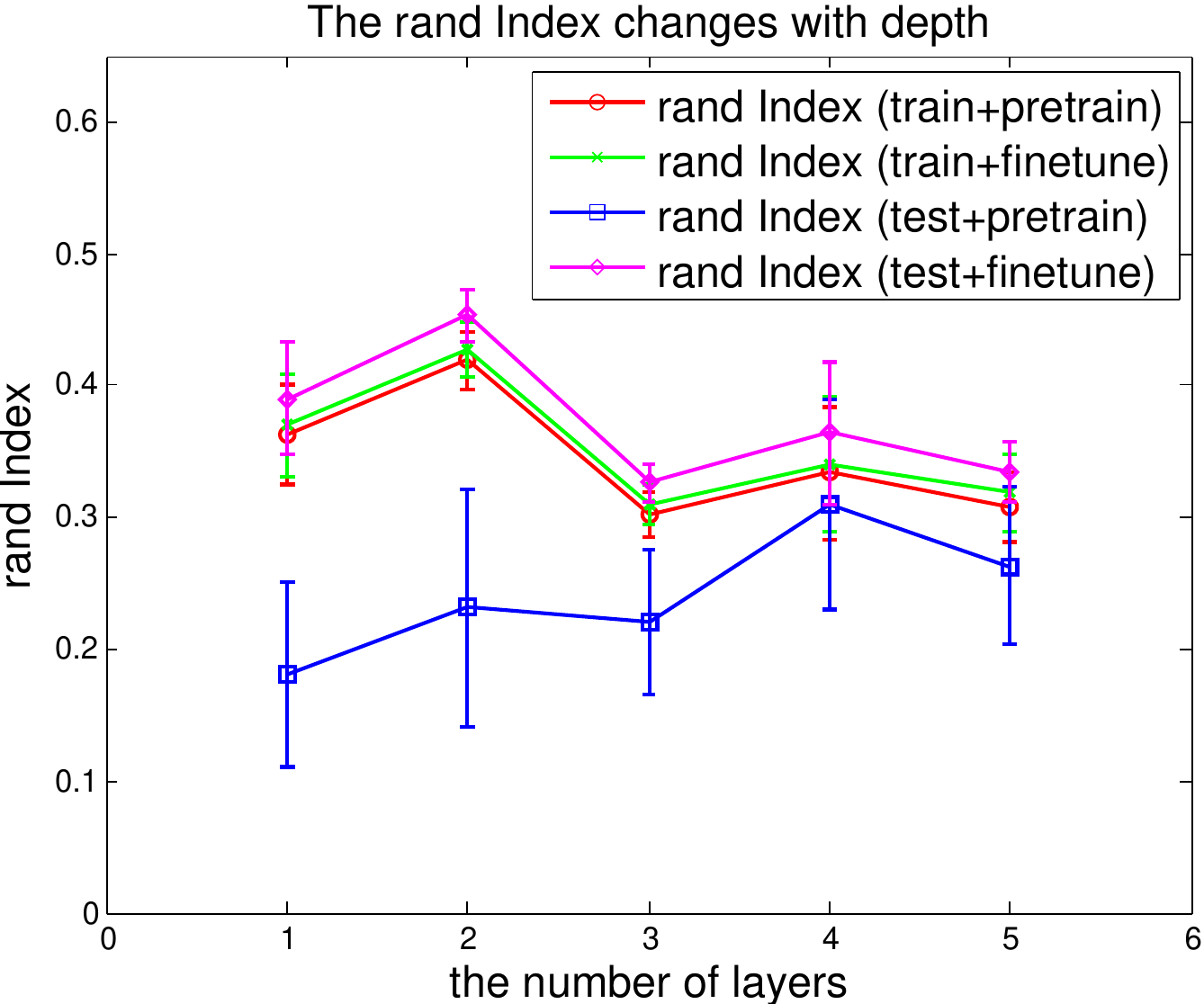}  \\ 
     (a) & (b)
     \end{tabular}
  \end{center}
  \caption{How the dimensionality and structural depth influence performance on MNIST dataset. (a) how the Rand Index changes with the encoded data dimension; (b) how the Rand Index changes with the depth of deep structures. It demonstrates the fine-tuning process is helpful to improve clustering performance. It also shows that complex deep structures cannot improve clustering accuracy. }
\label{fig:dim}
\end{figure*}

{\bf Clustering on MNIST dataset}: The MNIST dataset\footnote{\url{http://yann.lecun.com/exdb/mnist/}} consists of $28\times28$-size images of handwriting digits from $0$ through $9$ with a training set of 60,000 examples and a test set of 10,000 examples, and has been widely used to test character recognition methods. In the experiment, we randomly sample 5000 images from the training sets for parameter learning and 1000 examples from the testing sets to test our model. 
After learning the features with DBN in the pre-training stage, we used NMMC for clustering, with setting $\alpha = 4$, $\lambda = 15$ and $C = 0.001$. In the experiment, $\lambda$ plays a vital role on the final number of clusters. Higher $\lambda$, larger number of clusters generated. To make an fair comparison, we basically tuned parameters to keep the number of generated clusters close to the groundtruth in the training stage. For example, in the MNIST experiment, we keep it around 5 to 20 in the training set for both NMMC and DPM. The results from baselines such as k-means and GMM should be conceived as upper bound (specify the number of clusters $K=10$). 

The clustering performance of our method (DBN+NMMC) is shown in Table (\ref{tab:acc}), where ``pre-train" and ``fine-tune" indicate how the accuracy changes before and after the fine-tuning process for the same parameter setting on the same dataset. The results with 2-layer DBN in Table (\ref{tab:acc}) demonstrate that our method significantly outperforms baselines. It also shows that fine-tuning process can greatly improve accuracy, especially on the testing data. In Table (\ref{tab:acc}), we think the largest train/test difference for the least complex model is caused by biases between before and after finetuning. In other words, the fine-tuning step can learn better biases via classification RBM and improve testing performance. We also visualize how the weights change before and after the fine-tuning process in Fig. (\ref{fig:weights}). 

We also evaluate how the depth and dimensionality of deep structures influence clustering accuracy. Fig. \ref{fig:dim}(a) shows how adjusted Rand Index changes with the number of dimensions for 1-layer DBN (or RBM), and it demonstrates that higher dimensionality does not mean higher performance. In Fig. \ref{fig:dim}(a), we can see fine-tuning severely hurt performance on the training set on higher dimension coding space, we guess it is caused by overfitting problem in the complex model. In other words, the wrong clustering prediction will deteriorate the clustering performance even further through fine-tuning. That makes sense because we treat the wrong labeling as the correct one in the fine-tuning stage. It also verifies that it is reasonable by just fine-tuning the model in the top layer, instead of the whole network, with the purpose to reduce the overfitting problem. Fig. \ref{fig:dim}(b) shows that given the 100 hidden nodes in the top layer, how the performance changes with the depth of DBN structure. It seems that the deeper complex model cannot guarantee better performance.% \cite{Larochelle09}. 

\begin{figure}[h!]
  \begin{center}
     \begin{tabular}{cc}
     \includegraphics[trim = 35mm 83mm 42mm 83mm, clip, width=6.8cm]{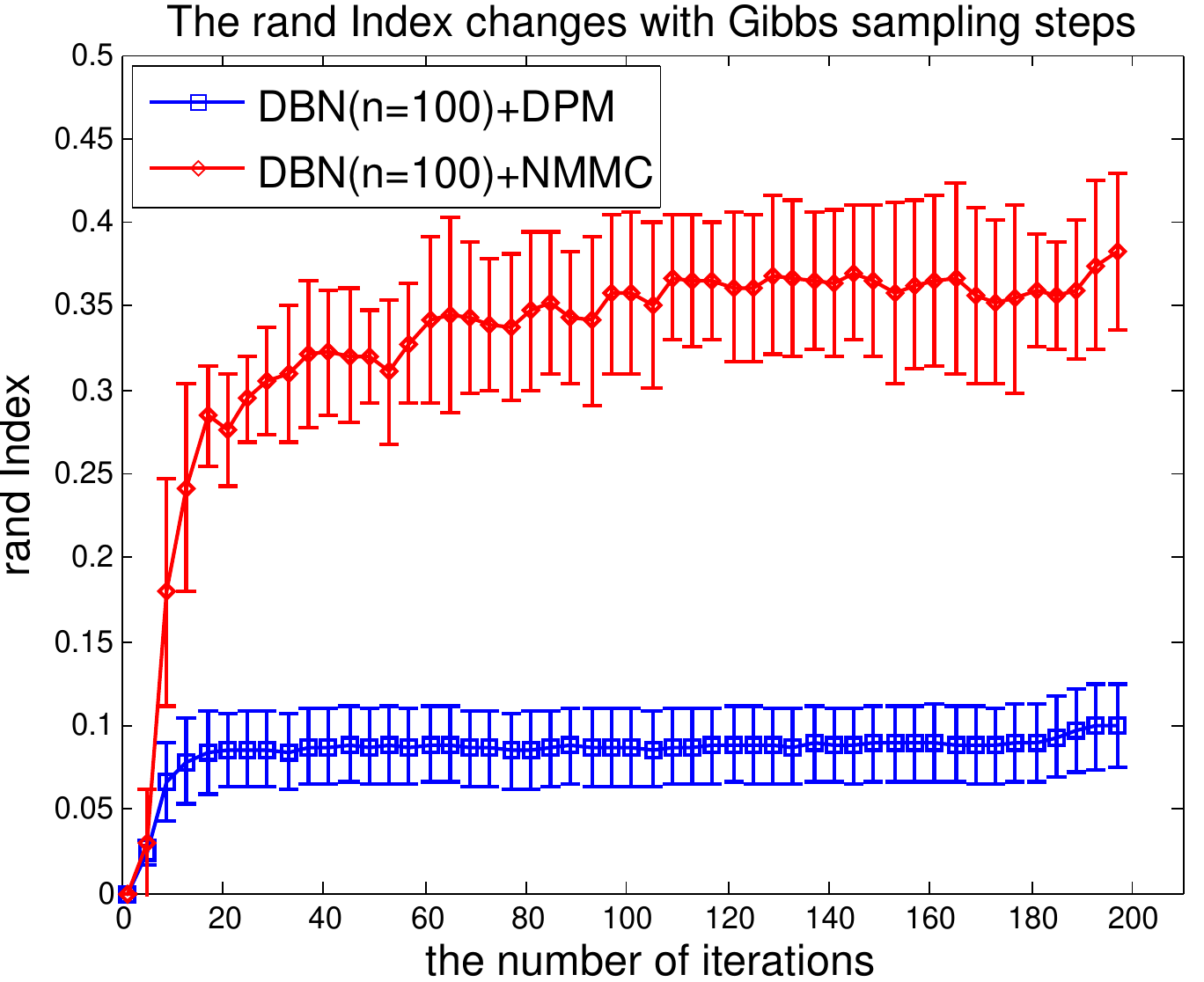} &
     \includegraphics[trim = 35mm 83mm 43mm 83mm, clip, width=6.8cm]{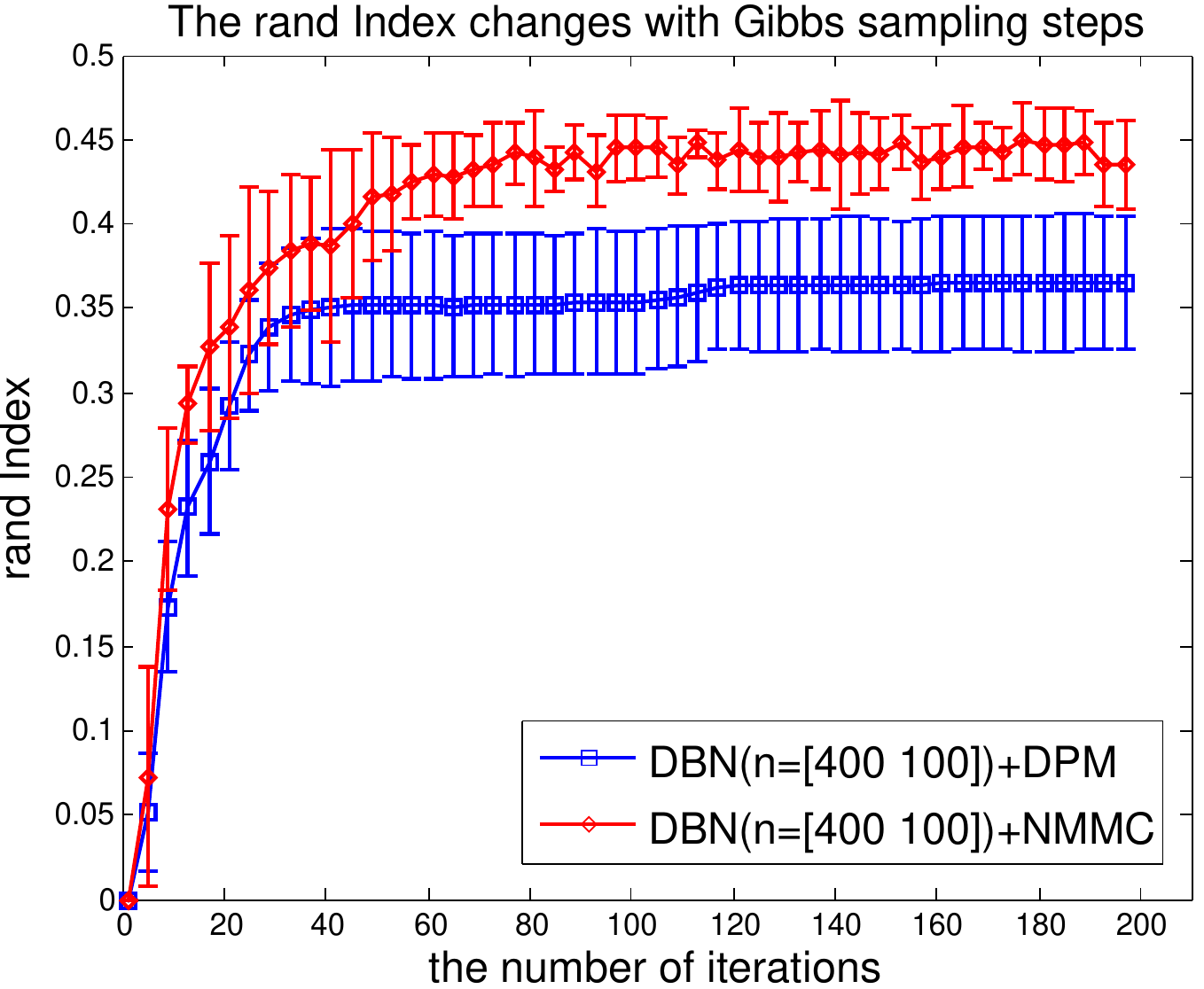}  \\ 
     (a) & (b)
     \end{tabular}
  \end{center}
  \caption{The performance comparison between DPM and NMMC on the MNIST dataset with the same DBN structure for feature learning. (a) it is a 1-layer DBN (or RBM) with the number of hidden nodes $n =100$; (b) it is a 2-layers DBN, with $n = [400, 100]$ for each layer. It demonstrates that with the same DBN for feature learning, NMMC outperforms DPM remarkably.}
   \label{fig:compdmp}
\end{figure}

\begin{figure}[h!]
  \begin{center}
     \begin{tabular}{cc}
     \includegraphics[trim = 34mm 83mm 40mm 83mm, clip, width=6.8cm]{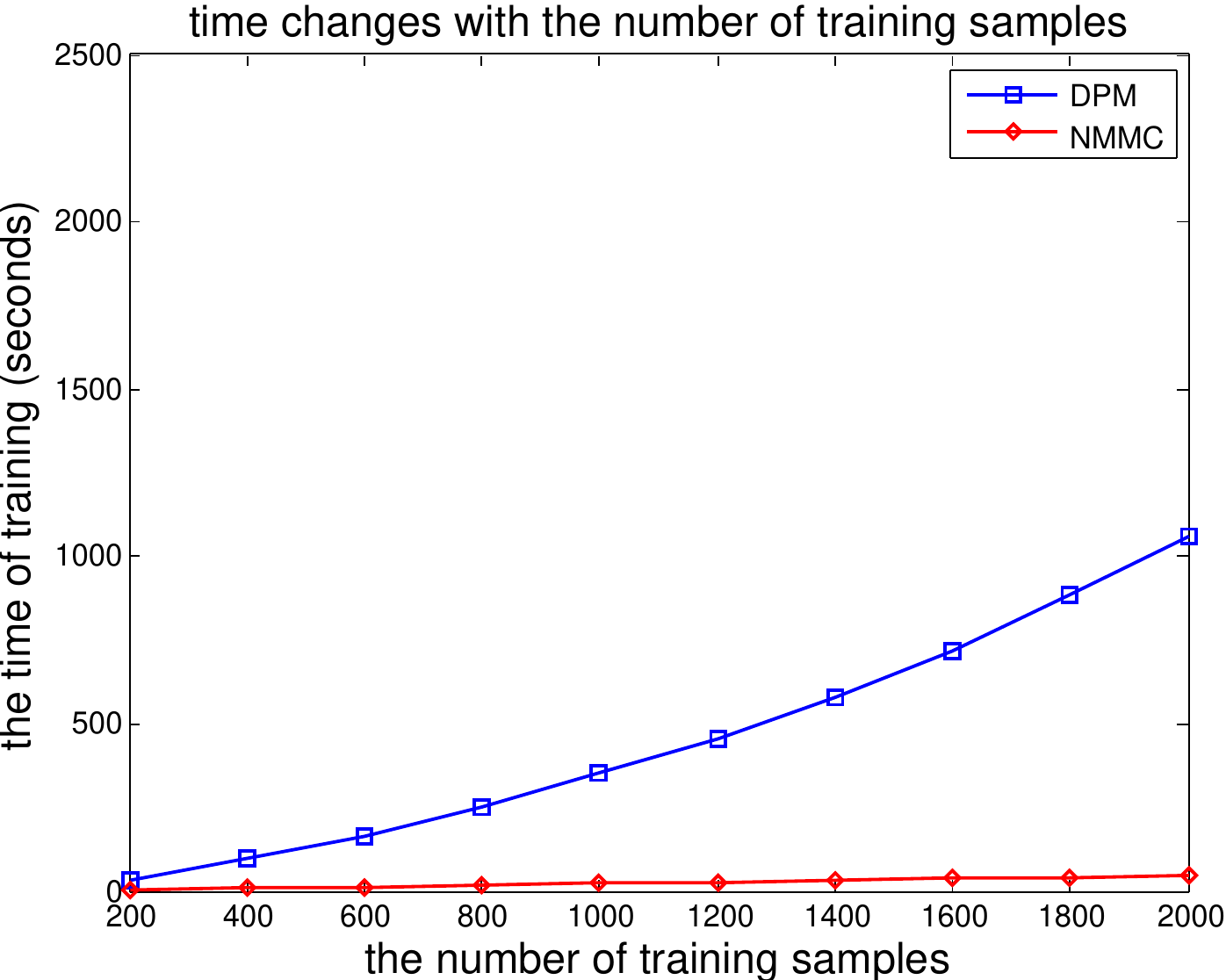} &
     \includegraphics[trim = 34mm 83mm 41mm 83mm, clip, width=6.8cm]{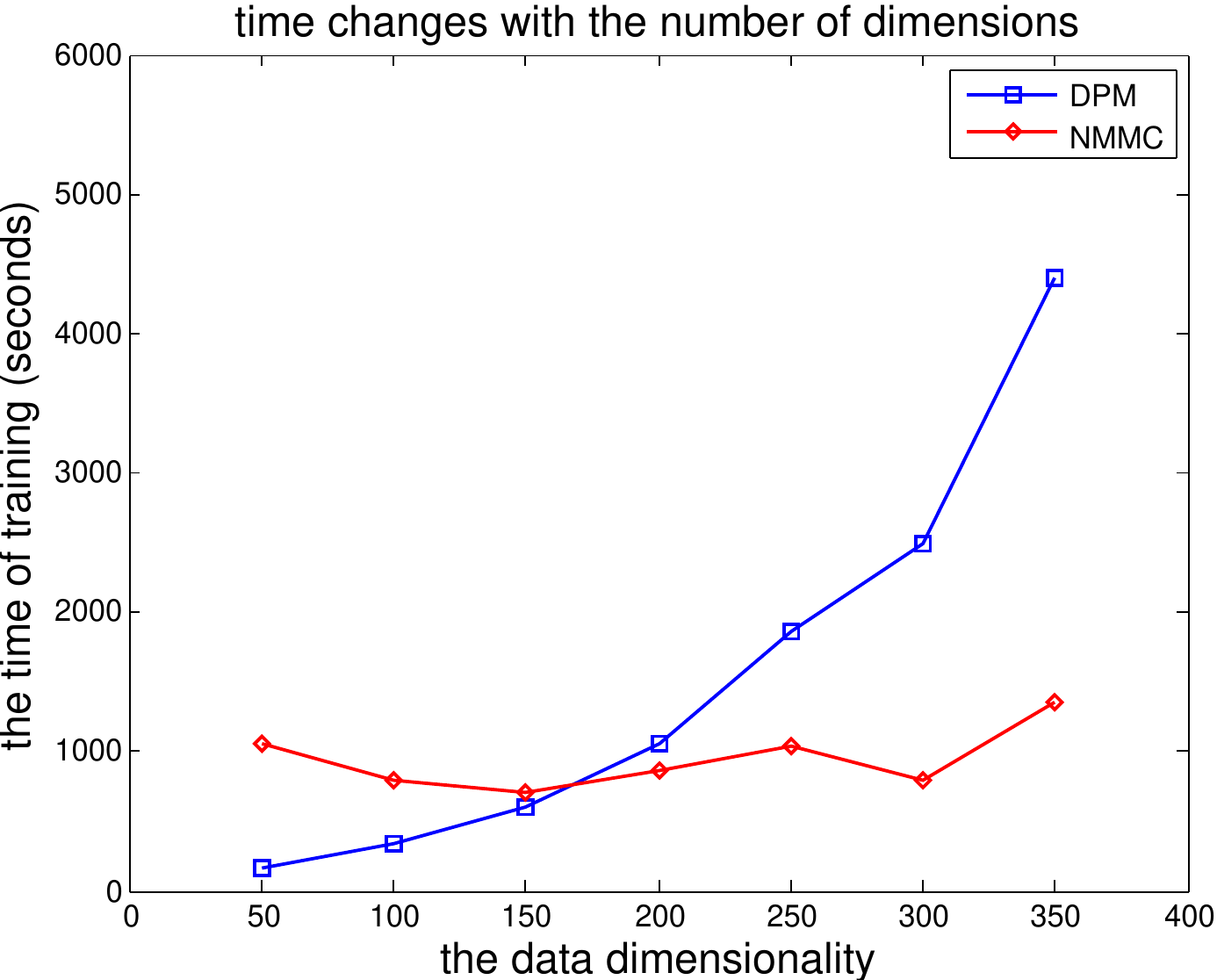}  \\ 
     (a) & (b)
     \end{tabular}
  \end{center}
  \caption{The complexity comparison between DPM and NMMC in the data projection space. (a) shows how the time complexity varies with the number of training data on the MNIST data set, under the 1-layer DBN with 100 hidden nodes; (b) shows how the time complexity changes with the number of hidden nodes on the 20 newsgroup dataset, under the 1-layer DBN. It shows that our method is more efficient than DPM on the data clustering.}
   \label{fig:comptime}
\end{figure}

To verify whether our NMMC is effective for data clustering and model selection, we also compare our NMMC to DPM given the same DBN for feature learning. %Fig. (\ref{fig:compdmp})  shows that our NMCC can always converge after 100 iterations. 
The results in Fig. (\ref{fig:compdmp}) demonstrates that NMMC outperforms DPM significantly and also shows that our NMMC can always converge after 100 iterations. The time complexity comparison between our method and DPM is shown in Fig. \ref{fig:comptime} in the DBN projection space. It shows that our method is significantly efficient, compared to DPM. To manifest how effective our method is, we also show the upper bound DBN+GMM, with 2 layers $n =[400, 100]$ in Table (\ref{tab:acc}). It shows that features learned with DBN are helpful for clustering, compared to raw data. It also shows that our method yields better clustering results than the upper bound. 
%\vspace{-0.75\skip\footins}
%\renewcommand{\footnoterule}{}
%\end{minipage}
\begin{table*}[t!]
%\centering
%\smallskip \begin{minipage}{14.0cm}
\centering
\resizebox{\textwidth}{!}{
\begin{tabular}{lcccc}
\hline
%\multirow{2}{*}{Model} & \multicolumn{3}{c}{Error rate (\%)} \\\cline{2-4}
%& Rand Index & F-value & V-value \\
\multirow{2}{*}{Model} & \multicolumn{2}{c}{rand Index}  &  \multicolumn{2}{c}{F-value} \\
\cline{2-5}
 & train & test   & train & test \\
\hline
DBN+NMMC (pre-train, $n =100$) & $0.363 \pm 0.038$  & $0.181 \pm 0.07$   & $0.442\pm 0.032$& $0.285\pm 0.063$ \\ 
DBN+NMMC (fine-tune, $n =100$)  & $0.371 \pm 0.039$   & $0.392 \pm 0.043$  & $0.447\pm 0.033$ & $0.467\pm 0.036$\\ 
%DBN+NMMC (pre-train, test, $n =100$)  & $0.181 \pm 0.07$ \\ 
%DBN+NMMC(fine-tune, test, $n =100$)  & $0.392 \pm 0.043$ \\ 
\hline
DBN+NMMC (pre-train, $n =[400, 100]$) & $0.419\pm 0.022$ & $0.232 \pm 0.09$   & $0.483\pm0.02 $ & $0.319\pm0.07$\\ %RBM ($\eta = 0.0005$, $n = 1000$) & 24.9\\
DBN+NMMC (fine-tune, $n =[400,100]$)  & ${\bf 0.428} \pm 0.021$ & ${\bf 0.453} \pm 0.02$ & ${\bf 0.492}\pm0.02$ & ${\bf 0.513}\pm0.016$\\ 
%DBN+NMMC (pre-train, test,  $n =[400,100]$)  & $0.232 \pm 0.09$ \\ 
%DBN+NMMC (fine-tune, test, $n =[400,100]$)  & $0.453 \pm 0.02$ \\ 
\hline
DBN+NMMC (pre-train,  $n =[400,400,100]$)  & $0.302 \pm 0.017$ & $0.218 \pm 0.055$ & $0.394\pm0.014$ & $0.317\pm0.046$\\ 
DBN+NMMC (fine-tune, $n =[400,400,100]$)  & $0.309 \pm 0.015$ & $0.326 \pm 0.015$ & $0.40\pm0.012$ & $0.415\pm0.02$\\ 
%DBN+NMMC (pre-train, test,  $n =[400,400,100]$)  & $0.218 \pm 0.055$ \\ 
%DBN+NMMC (fine-tune, test, $n =[400,400,100]$)  & $0.326 \pm 0.015$ \\ 
\hline
DBN+NMMC (pre-train, $n =[400,300,200,100]$)  & $0.334 \pm 0.05$ & $0.31 \pm 0.08$ & $0.423\pm0.04$ & $0.40\pm0.07$\\ 
DBN+NMMC (fine-tune, $n =[400,300,200,100]$)  & $0.34 \pm 0.051$ & $0.364 \pm 0.054$ & $0.433\pm 0.04$ & $0.45\pm0.045$\\ 
%DBN+NMMC (pre-train, test,  $n =[400,300,200,100]$)  & $0.31 \pm 0.08$ \\ 
%DBN+NMMC (fine-tune, test, $n =[400,300,200,100]$)  & $0.364 \pm 0.054$ \\ 
\hline
PCA+NMMC ($n=100$) & $0.381\pm 0.02$ & $0.251 \pm0.022$ & $0.452\pm0.02$ & $0.353\pm0.02$ \\
IMRBM \cite{Nair08} ($n =100$, $K=10$) & $0.13 \pm 0.04$ & $0.10 \pm 0.03$ & $0.23\pm 0.02$& $0.22\pm0.02$\\
%IMRBM \cite{Nair08} (test, $n =100$, $K =10$) & $0.10 \pm 0.03$ \\
\hline\hline
k-means ($K =10$)  & $0.356\pm0.029$ & $0.367\pm0.03$ & $0.446\pm 0.026$& $0.451\pm 0.026$\\
%k-means (test, $K =10$)  & $0.367\pm0.03$ \\
GMM ($K=10$) & $0.356\pm0.029$ & $0.394\pm0.04$ &$0.446\pm 0.025$& $0.465\pm 0.026$\\
%GMM (test, $K=10$) & $0.394\pm0.04$ \\
Spectral Clustering ($K=10$) & $0.354\pm0.057$ & $ 0.359 +0.035$ & $0.423\pm 0.045$ & $0.423\pm 0.03$\\
DBN + kmeans ($K=10$) & $0.411\pm0.016$ & $0.316\pm0.027$ & $0.473\pm 0.015$ &  $0.401\pm 0.019$\\
DBN + GMM ($K=10$) & ${\bf 0.411}\pm0.016$ & ${\bf 0.406}\pm0.022$ & ${\bf 0.473}\pm 0.015$ &  ${\bf 0.467}\pm 0.024$\\
%GMM (test, $K=10$) & $0.394\pm0.04$ \\
%DBN + Spectral Clustering ($K=10$) & $0.396\pm0.024$ & $ {\bf 0.432} +0.03$ & $0.458\pm0.02$ & ${\bf 0.510}\pm0.027$\\
\hline
\end{tabular}
}
\caption{The experimental comparison on MNIST dataset, where ``train'' means the training data, ``test'' indicates the testing data, $n$ specifies the number of hidden variables for each layer (for example, $n =[400, 100]$ indicates DBN has two layers, the first layer has 400 hidden nodes, and the second layer has 100 hidden nodes). For PCA+NMMC, we first use PCA project the data into 100 dimensions, then perform NMMC for clustering. It demonstrates that the fine-tuning process in our model can improve clustering performance greatly, and our method (DBN+NMMC) beats the baselines remarkably when $n =[400, 100]$. }%We compare the performances between our method and other baselines. It demonstrates that our method (DBN+NMMC) improves clustering accuracy. }
\label{tab:acc}
\end{table*}

{\bf Clustering on 20 newsgroup}:
We also evaluated our model on 20 newsgroup datasets for document categorization. This document dataset has 20 categories, which has been widely used in text categorization and document classification. In the experiment, we tested our model on the binary version of the 20 newsgroup dataset\footnote{\url{http://www.cs.toronto.edu/~larocheh/public/datasets/20newsgroups/20newsgroups_{train,valid,test}_binary_5000_voc.txt}}. We used the training set for training and tested the model on the testing dataset. After we learned features in the DBN, we used NMMC for clustering, with setting $\alpha = 4$, $\lambda = 30$ and $C = 0.001$. To make an fair comparison, we basically took a similar setting as in the MNIST dataset, for both NMMC and DPM in order to generate the number of clusters which is comparable for both methods. % In the experiment, DPM generated around 50-80 clusters, and our method generated around 40-60 clusters. 
Baselines such as k-means and GMM should be thought of as upper bound because they need to specify the number of clusters $K=20$. 

\begin{figure}[h!]
  \begin{center}
     \begin{tabular}{cc}
     \includegraphics[trim = 35mm 83mm 42mm 83mm, clip, width=6.8cm]{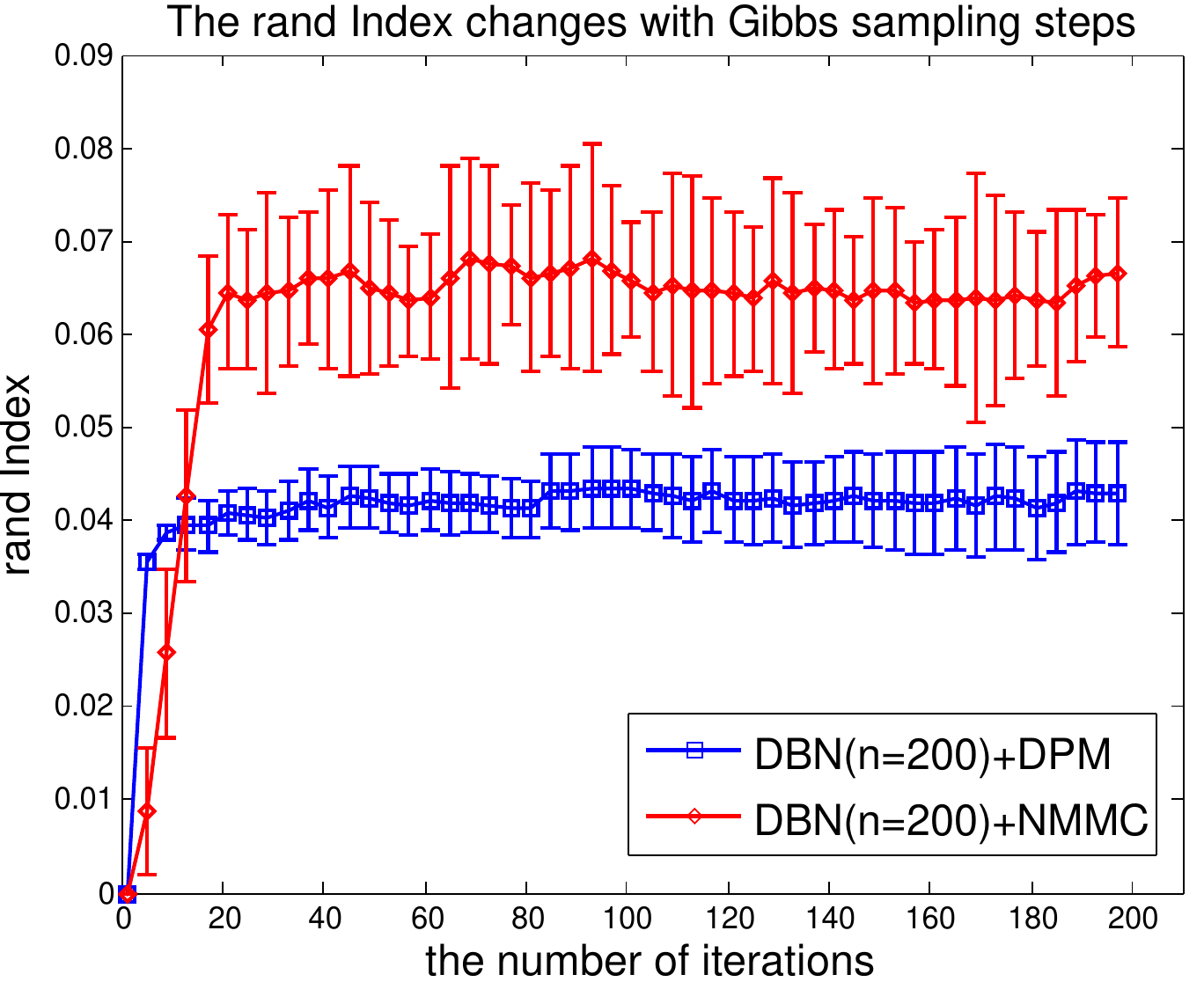} &
     \includegraphics[trim = 35mm 83mm 43mm 83mm, clip, width=6.8cm]{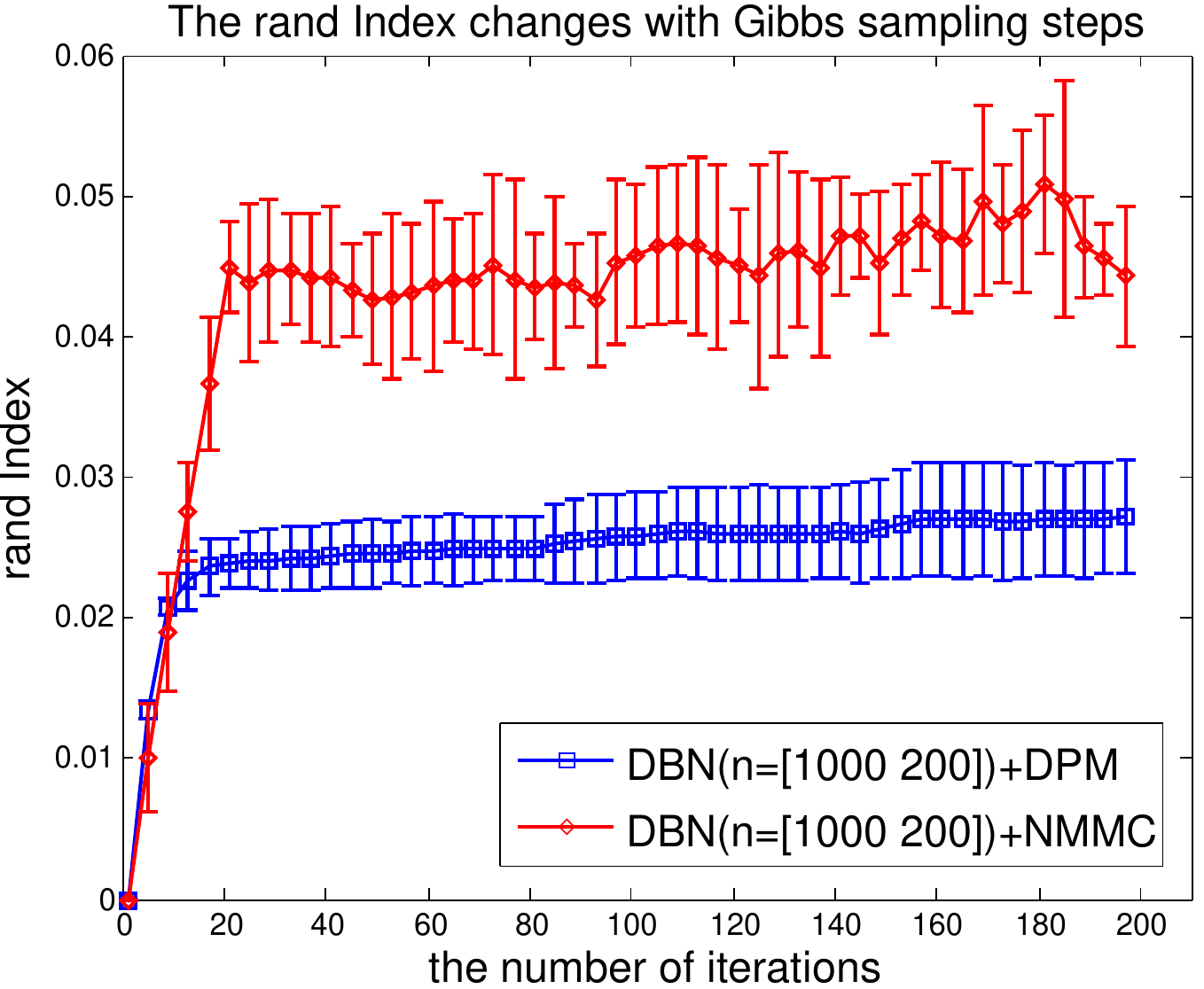}  \\ 
     (a) & (b)
     \end{tabular}
  \end{center}
  \caption{The performance comparison between DPM and NMMC with the same DBN structure for feature learning on 20 newsgroups. (a) it is a 1-layer DBN (or RBM) with the number of hidden nodes $n =200$; (b) it is a 2-layers DBN, with $n = [1000, 200]$ for each layer. It demonstrates that with the same DBN for feature learning, NMMC outperforms DPM remarkably.}
   \label{fig:compnews20}
\end{figure}

The clustering performance of our method (DBN+NMMC) on 20 newsgroups is shown in Table. (\ref{tab:acc_doc}). It also demonstrates that the fine-tuning process can greatly improve accuracy, especially on the testing data. Although our model cannot beat baselines on the training set, our model can achieve better evaluation performance on the testing set (better than GMM and k-means on the raw data clustering). To verify whether our NMMC is effective for data clustering and model selection, we also compare our NMMC to DPM given the same DBN for feature learning. The results in Fig. (\ref{fig:compnews20}) demonstrate that NMMC outperforms DPM remarkably. To test how time complexity changes with respect to the number of dimensions in the projected space, we tried different coding spaces and compared our method with DPM, with results shown in Fig. \ref{fig:comptime}. Again, it demonstrates our method is more efficient in practice.
%Fig. (\ref{fig:compnews20})  shows that our NMCC can always converge after 100 iterations. 
\begin{table*}[t!]
\centering
\resizebox{\textwidth}{!}{ 
\begin{tabular}{lcccc}
\hline
%\multirow{2}{*}{Model} & \multicolumn{3}{c}{Error rate (\%)} \\\cline{2-4}
%& Rand Index & F-value & V-value \\
\multirow{2}{*}{Model} & \multicolumn{2}{c}{rand Index} & \multicolumn{2}{c}{F-value}  \\
\cline{2-5}  
& train & test & train & test\\
\hline
DBN+NMMC (pre-train, $n =200$) & $0.059 \pm 0.02$ & $0.034 \pm 0.016$ & $0.131\pm0.017$ & $0.11\pm0.012$\\ 
DBN+NMMC (fine-tune, $n =200$)  & ${\bf 0.069} \pm 0.023$ & ${\bf 0.065} \pm 0.025$ &${\bf 0.142}\pm0.019$ & ${\bf 0.141}\pm0.02$\\  
%DBN+NMMC (pre-train, test, $n =200$)  & $0.034 \pm 0.016$ \\ 
%DBN+NMMC(fine-tune, test, $n =200$)  & $0.065 \pm 0.025$ \\ 
\hline
DBN+NMMC (pre-train, $n =[1000, 200]$) & $0.048\pm 0.014$ & $0.028 \pm 0.007$ & $0.109\pm0.005$  &$0.098\pm0.007$ \\ %RBM ($\eta = 0.0005$, $n = 1000$) & 24.9\\
DBN+NMMC (fine-tune, $n =[1000,200]$)  & $0.047 \pm 0.015$ & $0.043 \pm 0.013$ &$0.108\pm0.006$ & $0.104\pm0.004$ \\ 
%DBN+NMMC (pre-train, test,  $n =[400,100]$)  & $0.232 \pm 0.09$ \\ 
%DBN+NMMC (fine-tune, test, $n =[400,100]$)  & $0.453 \pm 0.02$ \\ 
%\hline
%DBN+NMMC (pre-train, $n =[400,400,100]$)  & $0.302 \pm 0.017$ & $0.218 \pm 0.055$ \\ 
%DBN+NMMC (fine-tune, $n =[400,400,100]$)  & $0.309 \pm 0.015$ & $0.326 \pm 0.015$ \\ 
%%DBN+NMMC (pre-train, test,  $n =[400,400,100]$)  & $0.218 \pm 0.055$ \\ 
%%DBN+NMMC (fine-tune, test, $n =[400,400,100]$)  & $0.326 \pm 0.015$ \\ 
%\hline
%DBN+NMMC (pre-train, $n =[400,300,200,100]$)  & $0.334 \pm 0.05$ & $0.31 \pm 0.08$  \\ 
%DBN+NMMC (fine-tune, $n =[400,300,200,100]$)  & $0.34 \pm 0.051$ & $0.364 \pm 0.054$ \\ 
%%DBN+NMMC (pre-train, test,  $n =[400,300,200,100]$)  & $0.31 \pm 0.08$ \\ 
%%DBN+NMMC (fine-tune, test, $n =[400,300,200,100]$)  & $0.364 \pm 0.054$ \\ 
\hline
PCA+NMMC ($n =200$) & $0.036 \pm 0.005 $& $0.016 \pm 0.012$ & $0.11\pm0.005$ & $0.087\pm0.010$\\
IMRBM \cite{Nair08} ($n =200$, $K=20$) & $0.015\pm 0.005$ & $0.013\pm0.002$ & $0.096\pm 0.004$& $0.093\pm 0.004$\\%$0.13 \pm 0.04$ & $0.10 \pm 0.03$ \\
%IMRBM \cite{Nair08} (test, $n =100$, $K =10$) & $0.10 \pm 0.03$ \\
\hline\hline
k-means ($K =20$)  & $0.075\pm0.02$ & $0.032\pm0.004$ & $0.140\pm0.019$& $0.109\pm0.016$\\
%k-means (test, $K =10$)  & $0.367\pm0.03$ \\
GMM ($K=20$) & $ 0.075\pm0.021$ & $0.051\pm0.006$ & $0.140\pm0.019$ & $0.114\pm0.016$\\
%GMM (test, $K=10$) & $0.394\pm0.04$ \\
Spectral Clustering ($K=20$) & $0.058\pm0.02$ & $0.061\pm 0.017$ & $0.126\pm0.013$& $0.129\pm0.006$\\
DBN + Kmeans ($K=20$) & $ 0.237 \pm0.007$ & $0.06\pm0.036$ &  $0.279\pm0.008$& $0.119\pm0.026$\\
DBN + GMM ($K=20$) & ${\bf 0.239}\pm0.009$ & ${\bf 0.125}\pm0.056$ & ${\bf 0.281}\pm 0.006$& ${\bf 0.185}\pm 0.045$\\
%GMM (test, $K=10$) & $0.394\pm0.04$ \\
%DBN + Spectral Clustering ($K=20$) & $0.265\pm0.008$ & $0.255\pm 0.005$ & $0.307\pm0.007$& $0.293\pm0.004$ \\
\hline
\end{tabular}
}
\caption{The experimental comparison on the 20 newsgroup dataset, where ``train'' means for training data, ``test'' indicates testing data. It demonstrates that the fine-tuning process in our model can improve clustering performance. We compare the performances between our method and other baselines. It demonstrates that our method (DBN+NMMC) yields clustering accuracy comparable to baselines, and performs better on the testing sets with 1-layer DBN.}
\label{tab:acc_doc}
\end{table*}

To sum up, our model can converge well after 100 iterations from the experiments above. Moreover, the fine-tuning process in our model can greatly improve the performance on the test sets. Thus, it also shows that the parameters learned with NMMC can be embedded well in the deep structures. 
\section*{Conclusion}
Clustering is an important problem in machine learning and its performance highly depends on data representation. And, how to adapt the model complexity with data also pose a challenge. In this paper, we propose a deep belief network with nonparametric maximum margin clustering. This approach is inspired by recent advances of deep learning for representation learning. As an unsupervised method, our model leverages deep learning for feature learning and dimension reduction. Moreover, our approach with nonparametric maximum margin clustering (NMMC) is a discriminative clustering method, which can adapt model size automatically when data grows. In addition, the fine-tuning process can incorporate NMMC well in the deep structures. %whose parameters can be learned online efficiently. 
Thus, our approach can learn features for clustering and infer model complexity in an unified framework. We currently use DBN \cite{hinton06a} instead of deep autoencoders \cite{Hinton06b} for fast feature learning because the latter is time-consuming for dimension reduction. In future work, we will explore deep autoencoders to learn better feature representation for clustering analysis. Another interesting topic to be explored is how to optimize the depth of  deep learning structures in order to improve clustering performance. % Acknowledgements should only appear in the accepted version. 
%\section*{Acknowledgments} 

\bibliographystyle{splncs03}
\bibliography{rbmbib}

\end{document}